\pdfoutput=1
\documentclass[11pt]{article}

\usepackage[final]{acl}

\usepackage{times}
\usepackage{latexsym}
\usepackage[T1]{fontenc}
\usepackage[utf8]{inputenc}
\usepackage{microtype}
\usepackage{inconsolata}
\usepackage{graphicx}
\usepackage{hyperref}       % hyperlinks
\usepackage{booktabs}       % professional-quality tables
\usepackage{amsfonts}       % blackboard math symbols
\usepackage{nicefrac}       % compact symbols for 1/2, etc.
\usepackage{xcolor}         % colors
\usepackage{amsmath}
\usepackage[ruled,vlined]{algorithm2e}
\usepackage{tikz}
\usepackage{pgfplots}
\usepackage{subcaption}
\usepackage{listings}
\usetikzlibrary{shapes, arrows.meta, positioning, fit}

\definecolor{amethyst}{rgb}{0.6, 0.4, 0.8}
\title{\textcolor{blue}{M}\textcolor{blue}{S}\textcolor{blue}{c}-SQL: \textcolor{blue}{M}ulti-\textcolor{blue}{S}ample \textcolor{blue}{C}ritiquing Small Language Models For Text-To-SQL Translation}

\author{
 \textbf{Satya Krishna Gorti},\ \
 \textbf{Ilan Gofman},\ \
 \textbf{Zhaoyan Liu},\ \
 \textbf{Jiapeng Wu},
\\
 \textbf{No\"el Vouitsis},\ \
 \textbf{Guangwei Yu},\ \
 \textbf{Jesse C. Cresswell},\ \
 \textbf{Rasa Hosseinzadeh},
\\
Layer 6 AI
\\
    \small{
    \texttt{\{satya, ilan, zhaoyan, paul, noel, guang, jesse, rasa\}@layer6.ai}
    }
 }

\begin{document}
\maketitle
\begin{abstract}
Text-to-SQL generation enables non-experts to interact with databases via natural language. Recent advances rely on large closed-source models like GPT-4 that present challenges in accessibility, privacy, and latency. To address these issues, we focus on developing small, efficient, and open-source text-to-SQL models. We demonstrate the benefits of sampling multiple candidate SQL generations and propose our method, MSc-SQL, to critique them using associated metadata. Our sample critiquing model evaluates multiple outputs simultaneously, achieving state-of-the-art performance compared to other open-source models while remaining competitive with larger models at a much lower cost. Full code can be found at \href{https://github.com/layer6ai-labs/msc-sql}{\texttt{github.com/layer6ai-labs/msc-sql}}.
\end{abstract}

\section{Introduction}\label{sec:intro}

Text-to-SQL generation is a rapidly growing area of natural language processing with significant real-world applications. It enables non-expert users to interact with databases using natural language queries, which are then automatically translated into SQL queries. This capability is widely applicable across various domains, including business intelligence, customer service automation, and data analysis. For example, it facilitates automating the feature selection process in tabular data prediction tasks which often involves complex aggregation queries over event histories. Furthermore, empowering translation models with database query access will allow for automated agentic workflows.

Recent advances in text-to-SQL generation have primarily leveraged closed-source models like GPT-4 \cite{openai2023gpt4} which, combined with advanced prompting techniques, have consistently achieved state-of-the-art performance on benchmarks such as Spider \cite{yu2018spider} and BIRD \cite{li2023birdbench}. However, the reliance on closed-source API-based models limits accessibility, transparency, and task adaptability, while presenting privacy concerns around data being sent to an API. All of these limitations underscore the need for efficient, open-source alternatives capable of competitive performance.

The inherent complexity of text-to-SQL tasks has contributed to proposed solutions becoming increasingly expensive. For instance, recent research \cite{wang2023mac, pourreza2024dts, talaei2024chess} demonstrates that decomposing complex tasks into specialized sub-tasks -- such as table prediction, SQL generation, and error correction -- improves overall performance. However, these composite approaches also increase the number of function calls to the model, leading to latency issues, especially when relying on larger models. This highlights the importance of developing smaller, task-specific models that maintain accuracy while being computationally efficient. Efforts such as DTS-SQL \cite{pourreza2024dts} and SFT CodeS-7B \cite{li2024codes} are among the few that try to address this need.

Our objective is to develop efficient methods for text-to-SQL generation that succeed with small and open-source models. We demonstrate that smaller language models (under 10B parameters) struggle to match the performance of their larger closed-source counterparts when relying solely on existing approaches, namely those that combine schema linking with SQL generation. We then show that this gap can be closed by sampling and running multiple SQL queries -- either from the same model or from an ensemble of models of similar size -- and comparing the results. By limiting the number of samples to two or three, this strategy balances between improving generation quality and maintaining computational efficiency. Recent works on other natural language generation tasks \cite{brown2024large, snell2024scaling, li2024more} support the idea that increased test-time compute can boost generation quality.

Sampling multiple SQL queries requires a way to judge the various candidates and then select the best one. Prior works on natural language generation with Large Language Models (LLMs) have employed techniques like training a ranker to evaluate the quality of generated samples \cite{li2022making}, or using reward models to guide the selection process \cite{ouyang2022training, rafailov2023dpo}. However, in the more specialized context of text-to-SQL generation we propose a distinct approach: training a sample-critiquing model that simultaneously considers multiple generations, their corresponding execution results, and associated metadata to determine which of the generated SQL queries should be returned. Compared to analogous methods, our approach allows the model to better leverage comparative information and contextual cues. Our results show state-of-the-art performance among open-source models on popular text-to-SQL benchmarks, while also achieving competitive results against larger closed-source models albeit at a much lower cost.

\begin{figure*}[t]
    \centering
    \includegraphics[width=\textwidth]{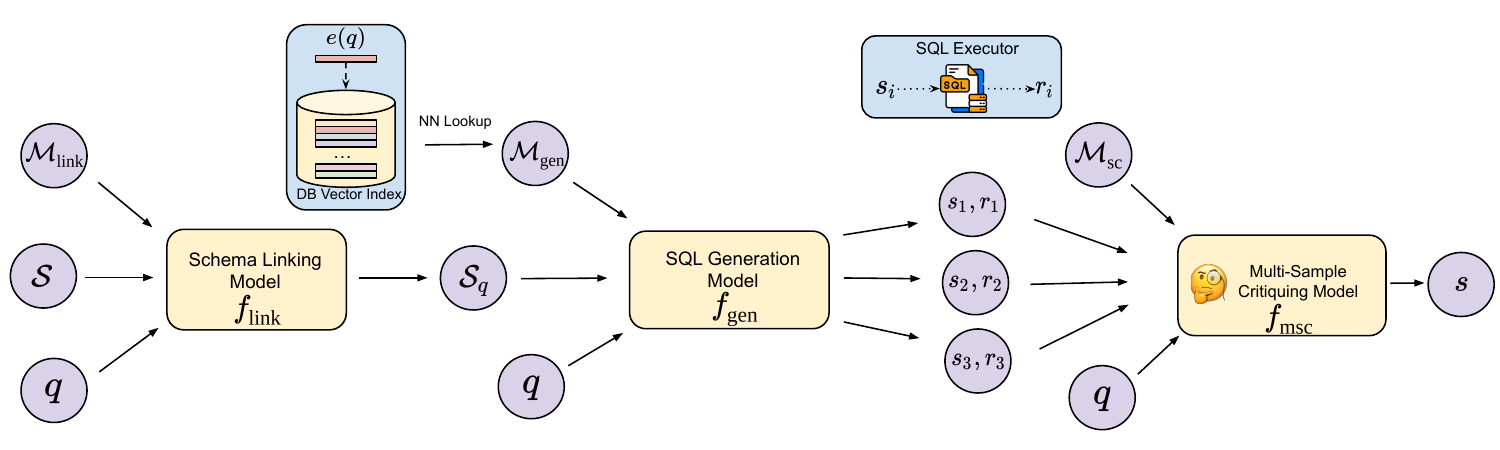}
    \caption{Starting with a natural language query $q$, database schema $\mathcal{S}$, and metadata $\mathcal{M}_{\text{link}}$, the schema linking model returns a subset $\mathcal{S}_{q}$ of tables which are necessary to answer $q$. Next, the SQL generation model adds metadata $\mathcal{M}_{\text{gen}}$ obtained through retrieval against an embedding of the query $e(q)$, and generates multiple possible SQL queries $s_i$. Finally, the multi-sample critiquing model comparatively evaluates the generations $s_i$ along with their execution results $r_i$ when run on the database, and then selects one as the final output $s$.}
    \label{fig:architecture}
\end{figure*}

\section{Related Work}\label{sec:related_work}

\textbf{Text-to-SQL Generation.} Early works in the space of text-to-SQL generation predominantly leveraged rule-based methods to parse natural language queries and map them to valid SQL statements \cite{popescu2003towards, popescu2004modern, li2014constructing}. Recently, LLMs such as GPT-4 \cite{openai2023gpt4} have facilitated this task by leveraging their strong textual priors to generate SQL queries \cite{li2023birdbench}. Due to the challenging nature of direct SQL generation, subsequent works instead decomposed the generation process into dedicated sub-tasks and prompted GPT-4 to solve each task sequentially \cite{wang2023mac, dong2023c3, pourreza2024din, chen2023teaching}. For example, MAC-SQL \cite{wang2023mac} defined three sub-components namely a Selector, a Decomposer, and a Refiner, where dedicated prompts were engineered for each component. Despite being state-of-the-art on several SQL generation benchmarks, the reliance of these methods on very large and closed-source models makes them inherently inefficient and expensive, and poses accessibility and privacy concerns. More recently, a few works have begun exploring the use of smaller open-source models (under 10 billion parameters) for SQL generation tasks \cite{pourreza2024dts, li2024codes} and have shown promising results. However, there remains a significant performance gap compared to several of the aforementioned GPT-4-based models. Our method MSc-SQL largely bridges this accuracy gap while remaining efficient and open-source.

\vspace{2pt}\noindent\textbf{Exploring Test-Time Computation.} On general LLM benchmarks, recent methods have sought to leverage additional test-time computation, such as planning, reasoning and problem deliberation to improve performance \cite{wei2022chain, yao2023react, besta2024graph, zelikman2024quiet, yao2024tree}. For example, Chain-of-Thought (CoT) prompting \cite{wei2022chain} forces the model to spend more tokens ``thinking'' about the problem before answering. Some methods have also explored repeated sampling as a way of using increased test-time computation to expand the space of generated solutions \cite{cobbe2021training, irvine2023rewarding, snell2024scaling, brown2024large, zhang2024accessing}. This expanded solution set is then filtered out using either rule based verifiers \cite{wang2022self}, such as picking the response that passes all the test-cases in the coding domain, or using other models to compare and then select the best sample among candidates \cite{cobbe2021training, lightman2023let}. We explore the latter direction in the domain of text-to-SQL generation.

\section{Towards Building an Efficient Text-to-SQL Pipeline}\label{sec:method}

\textbf{Problem Setup.} Text-to-SQL generation involves translating a natural language query \( q \) into a structured SQL query \( s \) that retrieves the desired information from a database. More formally, given a natural language question \( q \), the schema \( \mathcal{S} \) of a relational database \( \mathcal{D} \), and any extra metadata \( \mathcal{M} \) associated with the query such as extra evidence or few-shot examples from the database, the goal is to generate a valid SQL query \( s \) such that the execution of \( s \) on \( \mathcal{D} \) returns the correct answer to \( q \). In other words, the overall objective is to learn a model \( f: (q, \mathcal{S}, \mathcal{M}) \rightarrow s \). This problem is challenging on many fronts because of the diversity in which the natural language question can be expressed, the difficulty in inferring complex relational structures from a database schema, and the restriction of generating only valid SQL syntax for $s$. Due to these complexities, we divide the problem into three distinct modules which form a blueprint for an efficient text-to-SQL pipeline, depicted in \autoref{fig:architecture}. While the first two modules are commonly used, we make novel design recommendations in our blueprint. The third module described in \autoref{sec:sc} is original.

\subsection{Schema Linking}
Schema linking is the task of identifying the relevant tables and attributes within the schema \( \mathcal{S} \) that are necessary to construct a valid SQL query $s$ based on the natural language query \( q \). This step is critical because SQL queries typically involve only a subset of the available tables and attributes. Correctly identifying this subset ensures that the subsequent steps including SQL generation are focused on the most pertinent tables in the schema. Schema linking also reduces the input's length to help it fit in the limited context size of existing language models, while reducing costs in the quadratic Transformer attention \cite{vaswani2017attention} operation, and improving context utilization \cite{liu2024lost}. Given the schema \( \mathcal{S} \) containing information about each table and column type including the primary key and foreign key relationships, and corresponding metadata \( \mathcal{M}_\text{link} \), we denote our schema linking model as \(
f_{\text{link}}: (q, \mathcal{S}, \mathcal{M}_\text{link}) \rightarrow \mathcal{S}_q,
\) where \( \mathcal{S}_q \subseteq \mathcal{S} \) represents only the schemas of the tables that are predicted to be needed for answering \( q \). Our aim here is to maximize the recall of predicted tables; high recall is essential because missing even one necessary table immediately precludes complete and correct SQL generation. While emphasizing recall may introduce some false positives (i.e., irrelevant tables included in \( \mathcal{S}_q \)), our subsequent stages are designed to be robust to such inaccuracies.

\subsection{SQL Generation}

Given a reduced schema \( \mathcal{S}_q \), we then proceed to generate a SQL query \( s \) as $f_{\text{gen}}: (q, \mathcal{S}_q, \mathcal{M}_\text{gen}) \rightarrow s$, where $\mathcal{M}_\text{gen}$ is associated metadata. Although conceptually simple, generating a valid SQL query often requires knowledge of the formatting of column values, for instance when using a \texttt{WHERE} clause. To generate a correct SQL query $s$ for the example $q$ ``Which school is in California?'', $f_\text{gen}$ must know whether the state of California is represented as ``CA'', ``California'', or another variant thereof in the particular SQL database $\mathcal{D}$.

\vspace{2pt}\noindent\textbf{Contextual Retrieval through Few-Shot Examples.} We thus augment the SQL generation process by providing few-shot examples of the values in each column as additional metadata in $\mathcal{M}_\text{gen}$. By including this information, the model can better infer the correct SQL query, resolving potential ambiguities related to data representation in the database. For string columns it is especially important to provide information which is relevant to the query; hence we use the nearest neighbours of \( q \). The retrieval of few-shot examples is conducted through a similarity measure with an embedding of the input query concatenated with any additional evidence provided as metadata. A vector index on the entire database is constructed offline by indexing a fixed number of unique entries of every string type column. We specifically use \texttt{Alibaba-NLP/gte-large-en-v1.5} \cite{zhang2024mgte} as the embedding model. For other column types, we randomly sample few-shot examples.

\vspace{2pt}\noindent\textbf{Robust Training with Noisy Table Injection.} The schema linking stage may predict more tables than are strictly necessary in $\mathcal{S}_q$ because it aims to maximize recall. To combat this, we improve the robustness of generation to noise in $\mathcal{S}_q$ by having the SQL generation step learn to discard unnecessary and irrelevant tables. Starting with the ground-truth schema $\mathcal{S}_q^*$ from the training set containing only necessary tables needed for query $q$, we inject extra tables denoting the result by $\mathcal{S}_{q}^\dag$. We sample anywhere between 0 to 2 extra tables with a weighted probability, then train $f_\text{gen}$ on $\mathcal{S}_{q}^\dag$ rather than $\mathcal{S}_{q}^*$.

\vspace{2pt}\noindent\textbf{Training.} Starting with an open-source language model, we  fine-tune on tuples \( (q, \mathcal{S}_{q}^\dag, \mathcal{M}_{\text{gen}}, s) \) so that the model learns to generate syntactically correct and semantically valid SQL queries. The generation follows a sequence to sequence paradigm to maximize the likelihood of outputs,
\begin{equation}
    \mathcal{L}_{\text{gen}} = {-}\!\sum_{(q, \mathcal{S}_{q}^\dag, \mathcal{M}_{\text{gen}}, s)} \!\log P(s \mid q, \mathcal{S}_{q}^\dag, \mathcal{M}_{\text{gen}}),
\end{equation}
where $P$ represents the probability assigned to $s$ by the language model $f_\text{gen}$. Integrating both retrieval of contextual examples through $\mathcal{M}_{\text{gen}}$ and exposure to superfluous tables through $\mathcal{S}_{q}^\dag$ refines the model’s capability to discern relevant schema information, enhancing the accuracy of the generated SQL queries.

\section{MSc-SQL: Enhancing SQL Generation Using Multi-Sample Critiquing}\label{sec:sc}

As mentioned in \autoref{sec:intro}, our objective is two-fold. We propose to use only open-source language models for their improved accessibility, transparency, adaptability, and privacy. Second, we aim to achieve high-quality SQL generation using smaller language models to promote faster inference and reduce overall computational costs. However, small language models typically lag behind their larger counterparts in generation quality due to inherent limitations in their capacity. This is especially true in challenging tasks like SQL generation that require a deep understanding of the complex hierarchies present in an SQL statement.  We reconcile this gap by increasing the test-time computation budget used by smaller language models, sampling multiple generations to improve the likelihood of generating a correct SQL statement. In this work, we demonstrate that generating as few as three samples can yield highly competitive results on SQL benchmarks while maintaining the overall efficiency of the pipeline.

However, sampling multiple generations leaves us with the task of selecting the best possible generation among them. Prior works have used methods such as majority voting \cite{wang2022self}, use of a reward model \cite{christiano2017deep, cobbe2021training, lightman2023let,li2022making} or a combination of the two to pick the best generated candidate \cite{brown2024large} . This has shown promising results in other domains like math and coding tasks \cite{hendrycks2021measuring, li2022competition}. Other methods use pre-trained LLMs like GPT-4 to judge the quality of the sampled generations to select the best one \cite{zheng2023judging, lee2024mcs}. Such methods have the drawback of utilizing expensive closed models like GPT-4 to evaluate the best candidate. We mitigate this cost by developing an open model for the purpose of sample critiquing.

\subsection{Multi-Sample Critiquing in MSc-SQL} 

\begin{figure*}[t]
    \centering
    \includegraphics[width=\linewidth]{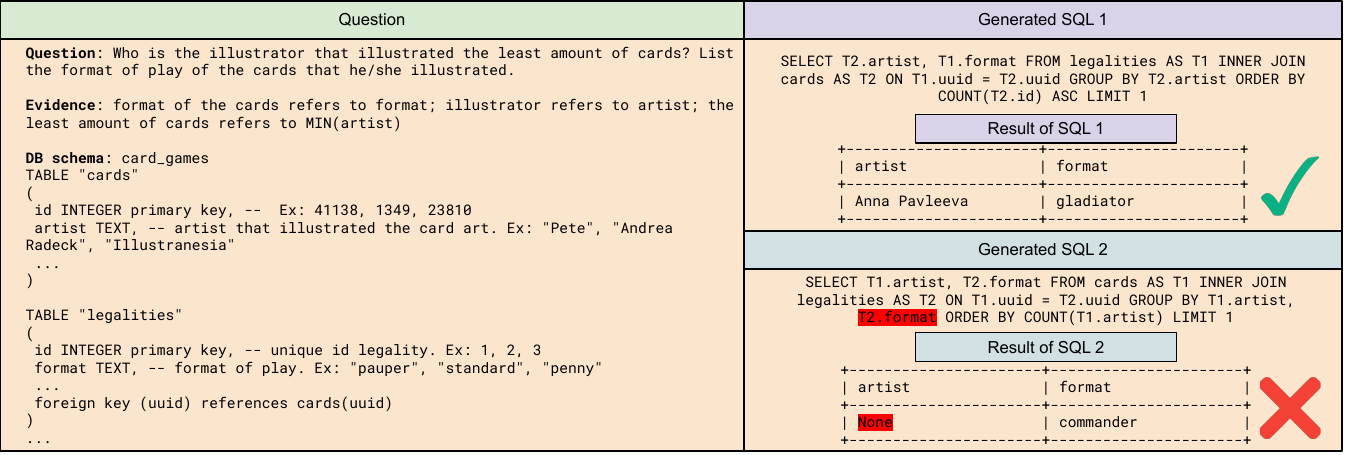}
    \caption{Two example queries sampled from our SQL generation model. Both are given to MSc-SQL for critiquing; one is correct and one is incorrect. Joint reasoning over both queries allows MSc-SQL to better capture the nuanced differences between them and thus select the correct query.}
    \label{fig:multi_critique}
    \vspace{-8pt}
\end{figure*}

Sample critiquing involves evaluating the inputs and the generated samples to determine the correctness of the generation. To facilitate this decision we can increase the contextual information provided to the model so it can make a more informed critique. To this end, we provide the model with not only the question, schema and the generated SQL queries, but also with the output of executing the SQL query and any resulting error messages. For example, if the query asks to find the average employee salary grouped by department, and a generated SQL query returns a single column of average salary without the grouping column, then the critiquing model should be able to identify the discrepancy and rule out the generated query. 
 
 We let \( s_i \) and \( r_i \) represent the candidate SQL queries and their respective execution results, and let $\mathcal{M}_{\text{sc}}$ denote any additional metadata needed for this stage. As a baseline, we first consider a basic critiquing model, using $f_{\text{isc}}$ to denote \textit{independent sample critiquing}, where $f_{\text{isc}}: (q, \mathcal{S}_{q}, (s_i, r_i), \mathcal{M}_{\text{sc}}) \rightarrow [0, 1]$. This model independently critiques the candidates $s_i$ which are then ranked according to the model's logits to output a single option. Although augmenting the input of $f_{\text{isc}}$ with $r_i$ improves the model's ability to select the correct SQL query, by independently ranking each SQL query the model still lacks the ability to jointly reason over multiple samples to consider relative differences among them; we expect that critiquing multiple samples at once should allow the model to consider subtle nuances that might make one query correct over another.

We therefore propose to simultaneously critique multiple generated samples (provided they fit within the maximum context size of the underlying model) to allow for a more comprehensive comparison. We illustrate an example in \autoref{fig:multi_critique} where we provide two generated SQL samples that MSc-SQL must choose between. The question asks to identify the illustrator with the least amount of cards, along with clarifying evidence on the `format' and `artist' columns. While the first generated query is correct, the second one wrongly groups the `artist' and `format' columns resulting in the value \textit{None} for the `artist' column (highlighted in red in the figure). Providing the model with both generations at once makes it easier to discern the correct response. We only illustrate two generated samples for clarity, but MSc-SQL generalizes to $n$ candidate samples as $f_{\text{msc}}: (q, \mathcal{S}, \{(s_i, r_i)\}_{i=1}^n, \mathcal{M}_{\text{sc}}) \rightarrow \{1, \dots, n\}$ where the output is now the index of the selected SQL sample. 

\vspace{2pt}\noindent\textbf{Training.} To facilitate training of our critiquing model and leverage the existing knowledge embedded in pre-trained open-source models, we model this task in a similar manner to next token prediction and simply predict the correct sample index as the next token.

\subsection{Inference} 
The end-to-end inference process is specified in Algorithm \ref{algo:inf}. In cases where the context size of the underlying model is not big enough to capture all the candidates, we can reduce the context size by employing pairwise comparisons on all pairs, a sliding window strategy \cite{sun2023chatgpt}, or a tournament-style voting mechanism where candidate pairs are arranged in a tournament bracket to reduce the number of comparisons required. Overall, we find that MSc-SQL yields highly competitive results, outperforming several methods that use proprietary LLMs like GPT-4, and stands as the best performing open-source model for text-to-SQL generation.

\begin{algorithm}[t]
\SetAlgoLined
\KwIn{Natural language query \( q \), schema \( \mathcal{S} \), metadata \( \mathcal{M}_\text{link}, \mathcal{M}_\text{gen}, \mathcal{M}_\text{sc} \)}
\KwOut{Generated SQL query \( s \)}
\BlankLine
\textbf{Step 1: Schema Linking}\\
Predict relevant schema subset: $\mathcal{S}_q = f_{\text{link}}(q, \mathcal{S}, \mathcal{M}_\text{link})$\\
\BlankLine
\textbf{Step 2: SQL Generation}\\
Retrieve nearest-neighbor examples and enrich metadata \( \mathcal{M}_\text{gen} \)\\
Generate candidates: $s_i = f_{\text{gen}}(q, \mathcal{S}_q, \mathcal{M}_\text{gen}), \quad i = 1, \dots, n$\\
\BlankLine
\textbf{Step 3: Sample Critiquing}\\
Execute candidates to obtain results \( r_i \)\\
Select the best candidate: $i^* = f_{\text{msc}}\left( q, \mathcal{S}_q, \{(s_i, r_i)\}_{i=1}^n, \mathcal{M}_\text{sc} \right)$\\
\textbf{Return} Selected SQL query \( s = s_{i^*} \)
\caption{Inference with MSc-SQL}
\label{algo:inf}
\end{algorithm}

\section{Experiments}\label{sec:experiments}

\begin{table*}[t]
 \small
  \centering
    \setlength{\tabcolsep}{1pt}
    \caption{Results comparing recent methods on the BIRD and Spider benchmarks. Params. refers to the number of model parameters; if multiple models are used we select the single largest.}
  \label{tab:combined}
  \begin{tabular}{l@{\hspace{1.5cm}}r}
   \hspace*{-1.0cm} 
    \begin{subtable}[b]{0.425\textwidth}
        \centering
        \begin{tabular}[t]{lccc}
            \toprule
            Method & Params. & Dev EX\%   \\
            \midrule
            \multicolumn{2}{c}{Closed Proprietary Models} \\
            \midrule
            Distillery+GPT-4o-finetune \scriptsize{\cite{maamari2024death}} & NA & \textbf{67.2} \\
            Distillery+Gemini-1.5-Pro \scriptsize{\cite{maamari2024death}} & NA & 60.5 \\
            CHESS+GPT-4 \scriptsize{\cite{talaei2024chess}} & NA & 65.0 \\
            MCS-SQL+GPT-4 \scriptsize{\citep{lee2024mcs}} & NA & 63.4  \\ 
            MAC-SQL+GPT-4 \scriptsize{\citep{wang2023mac}} & NA & 59.6 \\
            \midrule
            \multicolumn{2}{c}{Open Models} \\
            \midrule
            CHESS+Llama-3 \scriptsize{\cite{talaei2024chess}} & 70B & 61.5 \\
            Distillery+Llama-3 \scriptsize{\cite{maamari2024death}} & 405B & 59.9 \\
            SFT CodeS \scriptsize{\citep{li2024codes}} & 15B & 58.5  \\
            SFT CodeS \scriptsize{\citep{li2024codes}} & 7B & 57.2 \\
            DTS-SQL+DeepSeek \scriptsize{\citep{pourreza2024dts}} & 7B & 55.8  \\
            MSc-SQL (\textbf{Ours}) & 9B & \textbf{65.6} \\ 
            \bottomrule
        \end{tabular}
                \label{tab:bird_dev}
        \caption{BIRD Benchmark}
    \end{subtable}
    &
    \begin{subtable}[b]{0.4\textwidth}
      \centering
      \vspace{2.3mm}
        \begin{tabular}{lccc}
            \toprule
            Method & Params. & Test EX\%  \\
            \midrule
            \multicolumn{2}{c}{Closed Proprietary Models} \\
            \midrule
            CHESS+GPT-4 \scriptsize{\citep{talaei2024chess}} & NA & \textbf{87.2} \\
            DAIL-SQL+GPT-4 \scriptsize{\citep{gao2023texttosqlempoweredlargelanguage}} & NA & 86.6 \\
            DIN-SQL+GPT-4 \scriptsize{\citep{pourreza2024din}} & NA & 85.3 \\
            C3+ChatGPT \scriptsize{\citep{dong2023c3}} & NA  & 82.3 \\
            \midrule
            \multicolumn{2}{c}{Open Models} \\
            \midrule
            RESDSQL \scriptsize{\citep{li2023resdsql}} & 3B & 79.9 \\
            NatSQL+T5 \scriptsize{\cite{rai2023improving}} & 3B & 78.0 \\
            DTS-SQL+Mistral \scriptsize{\cite{pourreza2024dts}} & 7B & 77.0 \\
            MSc-SQL (\textbf{Ours}) & 9B & \textbf{84.7} \\ 
            \bottomrule
        \end{tabular}
              \label{tab:spider}
            \caption{Spider Benchmark}
        \end{subtable}
        \end{tabular}
        \vspace{-20pt}
\end{table*}

\textbf{Implementation Details.} For our experiments, we considered open-source language models with fewer than 10 billion parameters. Specifically, we used instruction-tuned variants of Mistral-7B-v0.3 \cite{jiang2023mistral}, Llama-3-8B \cite{dubey2024llama} and Gemma-2-9B \cite{team2024gemma} models. Fine-tuning was performed using QLoRA \cite{dettmers2024qlora}, with a LoRA rank of 32, a LoRA $\alpha$ of 128, and a dropout rate of 0.05. Fine-tuning was conducted with an effective batch size of 12, using a single NVIDIA A6000 Ada GPU with 48GB VRAM. The use of QLoRA allowed for memory efficient fine-tuning of these models without the need to update all the model parameters, in line with our efficiency goal.

\vspace{2pt}\noindent\textbf{Datasets and Metrics.} We utilized two primary datasets of text-to-SQL examples, Spider 1.0 \cite{yu2018spider} and BIRD \cite{li2023birdbench}, each serving distinct roles in evaluating the effectiveness of our approach to text-to-SQL generation. Spider 1.0 is a widely recognized dataset containing queries across 138 different domains spanning 200 databases. The BIg Bench for LaRge-scale Database Grounded Text-to-SQL Evaluation (BIRD) is a more comprehensive and challenging benchmark compared to Spider 1.0. BIRD contains over 12,751 unique question-SQL pairs on 95 big databases with a total size of 33.4 GB. All of the results in our paper are reported on Spider 1.0's test set and BIRD's Dev set.

Key evaluation metrics for text-to-SQL generation are: Execution Accuracy (EX), Exact Match (EM), and Validity and Efficiency Score (VES) \cite{yu2018spider, li2023birdbench}. EX measures the correctness of the SQL queries by checking if their execution results match the expected outcomes, making it a direct indicator of practical usability. EM assesses syntactic precision by comparing the generated SQL query against the reference query character by character; however, since different SQL queries can produce the same results, EM may punish functionally correct queries. VES evaluates both the correctness and computational efficiency of generated queries, which may be important in practical real-time systems.\footnote{Note that the run-time efficiency of the generated query is separate from the efficiency of the text-to-SQL method.} Still, generating correct queries is a challenging enough problem in its own right, so we prioritize EX and forgo evaluations of syntactic precision or efficiency.

\vspace{2pt}\noindent\textbf{Comparison with SoTA on BIRD and Spider 1.0 Datasets.} In \autoref{tab:combined} we provide detailed evaluation of our model's performance against the current state-of-the-art (SoTA) methods across both major benchmarks The results are grouped under two headings to emphasize the performance disparity between closed proprietary models and open models. Our method, MSc-SQL which integrates multi-sample critiquing, demonstrates impressive performance improvements across the two benchmarks compared to existing methods using open models. MSc-SQL critiques one sample from each of a fine-tuned Mistral-7B, Llama-8B, and Gemma-8B model to select the best candidate. Although, it only uses models under 10 billion parameters, our overall pipeline achieves a competitive score often outperforming methods that use proprietary language models such as Gemini-Pro \cite{reid2024gemini} and GPT-4 \cite{openai2023gpt4}, while also maintaining a significant advantage in inference speed. Compared to methods using open models, MSc-SQL outperforms the existing methods by a significant margin of 4.18 percentage points on BIRD.

\begin{table}[t]
        \centering
        \small
        \setlength{\tabcolsep}{4pt}
        \caption{Ablations of our method on BIRD. Superscripts $M$, $L$, and $G$ refer to the fine-tuned versions of Mistral, Llama-3, and Gemma-2 models respectively.}
        \vspace{-2pt}
        \label{tab:ablate}
        \begin{tabular}{ccc}
            \toprule
            Model Family & Params. & Dev EX\% \\
            \midrule
            $f_{\text{gen}}^M$ & 7B & 56.4 \\
            $f_{\text{gen}}^L$ & 8B & 54.1 \\
            $f_{\text{gen}}^G$ & 9B & 55.0 \\
            $f_{\text{link}}^M + f_{\text{gen}}^M$ & 7B & 61.3 \\
            $f_{\text{link}}^M + f_{\text{gen}}^L$ & 8B & 60.0 \\
            $f_{\text{link}}^M + f_{\text{gen}}^G$ & 9B & 60.8 \\
            $f_{\text{link}}^M \to \{ f_{\text{gen}}^M, f_{\text{gen}}^M \} \to f_{\text{msc}}^M $ & 7B & 64.0 \\
            $f_{\text{link}}^M \to \{ f_{\text{gen}}^L, f_{\text{gen}}^L \} \to f_{\text{msc}}^M $ & 8B & 63.5 \\
            $f_{\text{link}}^M \to \{ f_{\text{gen}}^G, f_{\text{gen}}^G \} \to f_{\text{msc}}^M $ & 9B & 62.7\\
            $f_{\text{link}}^M \to \{ f_{\text{gen}}^M, f_{\text{gen}}^M , f_{\text{gen}}^M \} \to f_{\text{msc}}^M $ & 7B & 64.8 \\
            $f_{\text{link}}^M \to \{ f_{\text{gen}}^L, f_{\text{gen}}^L , f_{\text{gen}}^L \} \to f_{\text{msc}}^M $ & 8B & 64.1 \\
            $f_{\text{link}}^M \to \{ f_{\text{gen}}^G, f_{\text{gen}}^G , f_{\text{gen}}^G \} \to f_{\text{msc}}^M $ & 9B & 62.9 \\
            $f_{\text{link}}^M \to \{ f_{\text{gen}}^M, f_{\text{gen}}^L , f_{\text{gen}}^G \} \to f_{\text{msc}}^M $ & 9B & 65.6 \\
            \bottomrule
        \end{tabular}
        \vspace{-8pt}
\end{table}

\vspace{2pt}\noindent\textbf{Ablations.} We show in Table \ref{tab:ablate} detailed ablation studies on different settings of our pipeline to understand how various parts of it contribute to the performance on the BIRD benchmark. We denote the underlying LLM using a superscript: $f^M$, $f^L$, $f^G$ denote the Mistral, Llama and Gemma models mentioned above. We first measure the effect of directly predicting SQL using $f_{\text{gen}}$, and then add schema linking before generation as $f_{\text{link}} \to f_{\text{gen}}$. Schema linking increases the overall accuracy by 5-6\% across generation models by removing redundant tables and improving focus. We then incorporate multi-sample critiquing by using either two or three samples from the underlying generation models and let the critiquing model $f_{\text{msc}}$ pick the most accurate SQL generation. Training such a model to critique the generations consistently increases the overall score across various settings as the sample size increases. Diversity of samples is also important. Sampling from each of a fine-tuned Mistral, Llama, and Gemma model results in the highest performance, achieving an accuracy of 65.6\% on the BIRD Dev set. We denote this setting as ``MSc-SQL'' in \autoref{tab:combined}.

We further perform analysis on the effect of using multiple generation models for sampling SQL outputs and contrast it with sampling multiple SQL outputs from a single generation model with non-zero temperature. Towards this, we train a number of different SQL generation models $f_{\text{gen}}$ from a base Mistral model with different random seeds, obtain one SQL sample from each model (with temperature set to zero), and use our critiquing model to pick the best candidate. We plot performance on the BIRD Dev set in \autoref{fig:sample_effect} and see that the improvement in accuracy saturates between three and five models. Limiting the number of samples from a smaller language model to three maintains the overall efficiency of the pipeline compared to using very large models. Since these models are trained using QLoRA, the memory footprint of using multiple versions is comparable to just loading the base model. 

\begin{figure}[t]
        \scriptsize
        \centering
        \caption{Effect of using different models to each create one sample for multi-sample critiquing. The generation models are all fine-tuned Mistral-7B models, but with different random seeds used during training.}
        \vspace{-2pt}
        \begin{tikzpicture}
            \begin{axis}[
                xlabel={No. of samples},
                ylabel={Dev Set EX (\%)},
                ymin=60, ymax=66,
                xmin=1, xmax=6,
                xtick={1,2,3,4,5,6},
                ytick={60, 61,62,63,64,65,66},
                grid=major,
                width=6cm, % Adjust width as needed
                height=5cm, % Adjust height as needed
                ]
                \addplot[mark=*] coordinates {
                    (1, 61.3)
                    (2, 64.01)
                    (3, 64.79)
                    (4, 64.86)
                    (5, 65.12)
                    (6, 64.91)
                };
            \end{axis}
        \end{tikzpicture}
        % \captionsetup{font=scriptsize}
        \label{fig:sample_effect}
        \vspace{-12pt}
\end{figure}

To quantify the effect of varying temperature while sampling from a single generation model $f_{\text{gen}}$, we fix the temperature $T$ to a value in $[0, 1]$, generate between one and five queries, and measure the performance of multi-sample critiquing on the BIRD Dev set with results shown in \autoref{tab:temp}. Of course, when $T$ is fixed to zero, sampling is deterministic and there is no benefit to critiquing. Increasing the number of samples and selecting the final output with $f_{\text{msc}}$ consistently increases the accuracy across different $T$ values. We note the best results with temperature $T=0.5$. 

\begin{table}[t]
    \small
    \centering
    \setlength{\tabcolsep}{2pt}
        
        \centering
        \caption{EX\% on the BIRD Dev set for varied temperatures $T$ and number of queries used for multi-sample critiquing. The underlying generation model is a fixed fine-tune of Mistral-7B.}
        \label{tab:temp}
        \vspace{-2pt}
        \begin{tabular}{lcccc}%${>{\raggedright\arraybackslash}m{4cm}cccc}
            \toprule
            Samples & $T=0$ & $T=0.1$ & $T=0.5$ & $T=1$ \\
            \midrule
            1 & 61.3 & 61.4 & 59.9 & 56.5 \\
            2 & 61.3 & 61.6 & 61.4 & 59.3 \\
            3 & 61.3 & 61.8 & 61.9 & 60.6 \\
            4 & 61.3 & 62.0 & 62.3 & 61.4 \\
            5 & 61.3 & 62.0 & 62.4 & 62.0 \\
            \bottomrule
        \end{tabular}
        \vspace{-8pt}
\end{table}

Based on these ablations, it is evident that sampling from diverse generation models and incorporating our multi-sample critiquing model increases the overall accuracy of the text-to-SQL pipeline. Compared to sampling from a single generation model with non-zero temperature, training different models from random initializations increases the likelihood of generating a correct query which can be picked out by multi-sample critiquing, and translates to higher performance. Both ablations highlight that diversity of samples is beneficial. Duplicate generated queries are not helpful in multi-sample critiquing, whereas generating diverse queries allows $f_\text{mcs}$ to contrast several approaches to the problem as well as execution results. Ultimately, only a single correct query is needed, but diverse samples provide more information to determine which may be correct.

\vspace{2pt}\noindent\textbf{Query Complexity}. In \autoref{tab:query_complexity} we compare how varying query complexity affects accuracy, using annotations from the BIRD benchmark. We see an overall improvement compared to most methods on all three categories. Importantly, we observe that MSc-SQL performs much better on simple queries compared to all the current methods, while CHESS+GPT-4 \cite{talaei2024chess} performs better on moderate and challenging categories. We believe this to due to the their use of GPT-4 to generate SQL queries which is expected to be much better at generating coherent complex sequences. Due to the plug-and-play nature of MSc-SQL where we can easily fine-tune and swap models in the pipeline, as smaller models improve in their ability to generate more complex sequences, the overall accuracy of our method can improve accordingly.

\begin{table}[t]
% \captionsetup{font=scriptsize}
    \footnotesize
    \centering
    \setlength{\tabcolsep}{2pt}
            \caption{Performance of methods as the complexity of BIRD queries is varied (EX\%).}
        \label{tab:query_complexity}
        \vspace{-2pt}
    % \hspace*{-0.7cm}
        \centering
        \begin{tabular}{lcccc}%${>{\raggedright\arraybackslash}m{4cm}cccc}
            \toprule
            & Simple & Moderate & Challenging & Overall \\
            \midrule
            GPT-4-turbo & 54.3 & 35.2 & 41.7 & 46.3 \\
            SFT CodeS-7B & 64.6 & 46.9 & 40.3 & 57.2 \\
            SFT CodeS-15B & 65.8 & 48.8 & 42.4 & 58.5 \\
            MAC-SQL+GPT-4 & 65.7 &  52.7 &  40.3 &  59.4 \\
            CHESS+GPT-4 & 65.4 & \textbf{64.8} & \textbf{58.3} & 64.6 \\
        
            MSc-SQL (\textbf{Ours}) & \textbf{72.0} & 58.0 & 49.0 & \textbf{65.6}\\
            \bottomrule
        \end{tabular}
        % \hspace{4mm}
        \vspace{-8pt}
\end{table}

\vspace{2pt}\noindent\textbf{Analyzing Multi-Sample Critiquing}. The data presented in \autoref{tab:sample_critiquing_comparison} highlights the effectiveness of the sample critiquing methodologies described in \autoref{sec:sc}. We compare these methods with an oracle critiquing model, $f_{\text{oracle}}$, that always chooses the correct query when there is at least one correctly generated SQL query in the available samples. With the oracle model, the pipeline's accuracy is 71.4\% on the BIRD Dev set, representing the ceiling in terms of the performance for a fixed generation model.

To show the efficacy of the multi-sample critiquing method with $f_{\text{msc}}$, we first compare two different critiquing models that process samples independently to generate a likelihood of correctness. One is an off-the-shelf high performing Llama-8B reward model taken from the RewardBench leaderboard \cite{lambert2024rewardbench}, \texttt{Skywork-Reward-Llama-3.1-8B}, that is tasked to predict the correctness of the generated output one at a time, with the highest ranked sample taken as the output. We refer to this model as $f_{\text{isc}}^1$. We also fine-tune a Llama-8B on the same dataset as we train $f_{\text{msc}}$, but with only a single query in context as described in section \ref{sec:sc}, and choose the highest ranked result. We refer to this as $f_{\text{isc}}^2$. Additionally, we evaluate a third approach based on self-consistency \cite{wang2022self}, which selects the most consistent output among the generated outputs, referred as $f_{\text{consistency}}$. We see that $f_{\text{isc}}^1$ performs worse than the baseline (i.e. no critiquing) which is likely due to the reward model not being specifically trained on SQL tasks. While we see an improvement of 1.6 p.p. from training $f_{\text{isc}}^2$ on SQL data. The self-consistency approach $f_{\text{consistency}}$ shows 1.3 p.p. improvements over the baseline, comparable to $f_{\text{isc}}^2$.  Multi-sample critiquing $f_{\text{msc}}$ outperforms all other selection methods by a large margin; an improvement of 4.3 p.p. over the baseline and 2.7 p.p. over $f_\text{isc}^2$, further demonstrating the benefits of critiquing multiple samples at once.

\begin{table}[t]
    \small
        \centering
            \setlength{\tabcolsep}{2pt}
                \caption{Measuring the effect of different sample critiquing techniques on BIRD.}\label{tab:sample_critiquing_comparison}
                \label{tab:sample_critiquing}
                        \vspace{-2pt}
        \begin{tabular}{lc}
        \toprule
        Method & Dev EX\% \\
        \midrule
        Baseline: $f_{\text{link}}^M \to f_{\text{gen}}^M$ & 61.3 \\
        $f_{\text{link}}^M \to  \{ f_{\text{gen}}^M, f_{\text{gen}}^L , f_{\text{gen}}^G \} \to f_{\text{oracle}}$ & 71.4 \\
        $f_{\text{link}}^M \to \{ f_{\text{gen}}^M, f_{\text{gen}}^L , f_{\text{gen}}^G \} \to f_{\text{isc}}^1 $ & 60.2 \\
        $f_{\text{link}}^M \to \{ f_{\text{gen}}^M, f_{\text{gen}}^L , f_{\text{gen}}^G \} \to f_{\text{isc}}^2 $ & 62.9  \\
        $f_{\text{link}}^M \to \{ f_{\text{gen}}^M, f_{\text{gen}}^L , f_{\text{gen}}^G \} \to f_{\text{consistency}} $ & 62.6 \\
        $f_{\text{link}}^M \to \{ f_{\text{gen}}^M, f_{\text{gen}}^L , f_{\text{gen}}^G \} \to f_{\text{msc}}^M $ & 65.6 \\
        \bottomrule
        \label{tab:msc_analysis}
        \end{tabular}
        \vspace{-8pt}
\end{table}

\section{Conclusion}\label{sec:conclusion}
We present an approach for text-to-SQL translation that leverages open-source language models to build an efficient and high performing method, and show comprehensive evaluations on two existing benchmarks along with analysis on various design choices made in our approach. We show that multi-sample critiquing is needed to address the limitations of smaller language models to compete with larger and proprietary counterparts. Our critiquing model is trained to leverage richer contextual information, including query execution results and errors to determine the best generation among candidate samples.

\newpage
\section{Limitations \& Risks}\label{sec:limitations}
While our approach performs competitively with most methods that use closed models, the choice to use smaller language models poses an inherent challenge on the capability of generating complex SQL statements. This was observed to be the case when analyzing responses qualitatively based on query complexity. We found that reliance on smaller language models necessitated the use of the critiquing step for adequate performance, which increases the complexity of the inference pipeline, even though the overall computational demand is lower than competing methods. We leave further investigation of the effects of critiquing larger open models, the effects of multi-sample critiquing on them, and the trade-offs around cost and performance to future work. 

There can also be risks that arise when relying on text-to-SQL models or more generally agentic workflows involving LLMs. Such models could inadvertently, or by way of an attack, output SQL queries that damage the databases which are accessible, such as by dropping tables. Our proposed method MSc-SQL is designed to automatically execute generated SQL code on the database and further process the outputs. Hence, if the generated code is damaging or malicious it could be automatically executed leading to harms. Any such workflow that automatically runs generated code should be implemented with guardrails to prevent permanent damage, for instance by backing up databases, and segregating the agentic computing environment from other production systems.

\bibliography{bib}

\appendix
\newpage
\onecolumn
\section{Appendix}
We specify example prompts and outputs used to train the schema linking, SQL generation and multi-sample critiquing models.
\subsection{Schema Linking}

\begin{lstlisting}
User: As an experienced and professional database administrator, your task is to analyze a user question and a database schema to provide relevant information. You are given an SQL Question, "Evidence" which is information that you need to use to solve the question, "DB schema" containing the database schema.

Identify and list all the relevant tables names from the DB schema based on the user question, database schema and evidence provided. Make sure you include all of them.

SQL Question: What is the highest eligible free rate for K-12 students in the schools in Alameda County?

Evidence: Eligible free rate for K-12 = `Free Meal Count (K-12)' / `Enrollment (K-12)'

DB schema: california_schools
TABLE "frpm"
(
 "CDSCode" TEXT primary key,
 "Academic Year" TEXT,
 "County Code" TEXT,
 "District Code" INTEGER,
 ...
 2013-14 CALPADS Fall 1 Certification Status INTEGER,
 foreign key (CDSCode) references schools(CDSCode)
)

TABLE "satscores"
(
 "cds" TEXT primary key,
 "rtype" TEXT,
 "sname" TEXT,
 "dname" TEXT,
 ...
 foreign key (cds) references schools(CDSCode)
)

TABLE "schools"
(
 "CDSCode" TEXT primary key,
 "NCESDist" TEXT,
 "NCESSchool" TEXT,
 "StatusType" TEXT,
 ...
 LastUpdate DATE,
)

Assistant:
```json
{
    "tables": ["frpm", "schools"]
}
```
\end{lstlisting}

\subsection{SQL Generation}

\begin{lstlisting}
User: You are given an SQL Question, "Evidence" which is information that you need to use to solve the question, 'DB schema' containing the database schema.
Think step by step and solve the question by coming up with the correct SQL statement that solves the question.

Important things to keep in mind:
1. Only use the tables that are required to solve the task.
2. Use the "evidence" and sample values and column descriptions given to you for reasoning.
3. Don't forget keywords such as DISTINCT, WHERE, GROUP BY, ORDER BY, LIMIT, etc. if needed.

SQL Question: What is the highest eligible free rate for K-12 students in the schools in Alameda County?

Evidence: Eligible free rate for K-12 = `Free Meal Count (K-12)` / `Enrollment (K-12)`

DB schema: california_schools
TABLE "frpm"
(
 "CDSCode" TEXT primary key, --  Example: "10621660114355", "10621176113575", "10621256113849", 
 "Academic Year" TEXT, --  Example: "2014-2015", 
 "County Name" TEXT, --  Example: "Alameda", "Contra Costa", "Santa Clara", 
 "District Name" TEXT, --  Example: "Alameda County Offic...", "San Francisco County...", "San Mateo County Off...", 
 ...
 foreign key (CDSCode) references schools(CDSCode)
)

TABLE "schools"
(
 "CDSCode" TEXT primary key, --  Example: "04755070000000", "04615490000000", "04746820000000", 
 "County" TEXT, -- County name Example: "Alameda", "Contra Costa", "Santa Clara", 
 "District" TEXT, --  Example: "Alameda County Offic...", "San Francisco County...", "San Mateo County Off...", 
 "School" TEXT, --  Example: "Alameda County Oppor...", "Alameda County Speci...", "Alameda City Unified...", 
 "City" TEXT, --  Example: "Alameda", "Santa Clara", "San Mateo", 
 ...
)

Reminder of the SQL Question: What is the highest eligible free rate for K-12 students in the schools in Alameda County?

Assistant: 
```json
{
    "sql": "SELECT `Free Meal Count (K-12)` / `Enrollment (K-12)` FROM frpm WHERE `County Name` = 'Alameda' ORDER BY (CAST(`Free Meal Count (K-12)` AS REAL) / `Enrollment (K-12)`) DESC LIMIT 1"
}
```
\end{lstlisting}

\subsection{Multi-Sample Critiquing}

\begin{lstlisting}
User: You are an SQL, database expert. A previous user was given a task of writing a SQL query given a question prompt. The user wrote 2 possible SQL queries. One is correct and one is wrong given the question. You task is to use the question and your expertise to decide which one is correct. Here is the question prompt: 

You are given an 'SQL Question', 'Evidence' which is information that you need to use to solve the question, 'DB schema' containing the database schema.
Think step by step and solve the question by coming up with the correct SQL statement that solves the question.

Important things to keep in mind:
1. Only use the tables that are required to solve the task.
2. Use the "evidence" and sample values and column descriptions given to you for reasoning.
3. Don't forget keywords such as DISTINCT, WHERE, GROUP BY, ORDER BY, LIMIT, etc. if needed.

SQL Question: Please list the phone numbers of the direct charter-funded schools that are opened after 2000/1/1.

Evidence: Charter schools refers to `Charter School (Y/N)` = 1 in the frpm

DB schema: Database: california_schools
TABLE "frpm"
(
 CDSCode TEXT primary key, --  Example: "10621661030675", "04755070433953", "10621661035831", 
 "Academic Year" TEXT, --  Example: "2014-2015", 
 "County Code" TEXT, --  Example: "33", "48", "49", 
 "District Code" INTEGER, --  Example: 10017, 31609, 31617, 
 ...
 foreign key (CDSCode) references schools(CDSCode)
)

TABLE "schools"
(
 CDSCode TEXT primary key, --  Example: "01316090000000", "04755070000000", "04755070433953", 
 NCESDist TEXT, -- This field represents the 7-digit National Center for Educational Statistics (NCES) school 
 ...
 LastUpdate DATE, --  Example: 2015-06-23, 2015-09-01, 2015-06-18, 
)


Reminder of the SQL Question: Please list the phone numbers of the direct charter-funded schools that are opened after 2000/1/1.


 The following are the two SQL queries written by the user along with the sample results they generated. One is correct and one is wrong. You need to decide which one is correct. 

1: SELECT DISTINCT T2.Phone, T1.CDSCode FROM frpm AS T1 INNER JOIN schools AS T2 ON T1.CDSCode = T2.CDSCode WHERE T1.`Charter School (Y/N)` = 1 AND T2.OpenDate > '2000-01-01' AND T1.FundingType = 'Directly funded'
Results of 1st SQL:
+----------------------+---------------------+
| Phone                | CDSCode             |
+----------------------+---------------------+
| (510) 596-8901       | 01100170109835      |
| (510) 563-1504       | 01100170118489      |
| (510) 146-7526       | 01100170130419      |
+----------------------+---------------------+
2: SELECT T2.Phone FROM frpm AS T1 INNER JOIN schools AS T2 ON T1.CDSCode = T2.CDSCode WHERE T1.`Charter Funding Type` = 'Directly funded' AND T2.OpenDate > '2000-01-01'
Results of 2nd SQL:
+----------------------+
| Phone                |
+----------------------+
| None                 |
| (510) 596-8901       |
| None                 |
| (510) 686-4131       |
| (510) 452-2063       |
+----------------------+

Provide the number of the correct SQL.

Assistant:
```json
{
    "correct_sql": "2"
}
```
\end{lstlisting}

\section{Latency and Computational Cost Analysis}

To evaluate the trade-off between latency and performance, we measured the inference time of our baseline model $f_{\text{gen}}^M$ and compared it with various configurations in Table \ref{tab:ablate}. All models are assumed to be preloaded into memory, and SQL generation is parallelized where no sequential dependencies exist. Our method incurs a nearly $1.7\times$ increase in latency compared to the standalone model, but this results in a performance gain of nearly 10\% on the BIRD Dev set, demonstrating an effective trade-off between inference speed and accuracy.

A direct latency comparison with other Text2SQL approaches in Table \ref{tab:combined} is infeasible due to proprietary models and the unavailability of open-source implementations. Instead, we estimate the computational cost in FLOPs for the strongest baseline by assuming a fixed context length of 2048 tokens per model invocation. We utilize the \texttt{calculate-flops.pytorch}\footnote{https://github.com/MrYxJ/calculate-flops.pytorch} library to compute total FLOPs across all function calls. The results are summarized in Table \ref{tab:latency}.

Compared to the CHESS baseline, which employs a 70B LLaMA model with multiple queries, our method requires over 10 times less compute while achieving superior performance. This highlights the efficiency of leveraging additional test-time computation with smaller models to outperform larger models while reducing computational costs.

\begin{table}[h]
\small
    \centering
    \begin{tabular}{lcccc}
        \toprule
        \textbf{Model Family} & \textbf{Params.} & \textbf{Dev EX\%} & \textbf{Latency (rel. to baseline)} & \textbf{FLOPs (TFLOPS)} \\
        \midrule
        $f_{\text{gen}}^M$ & 7B & 56.4 & 1.00 & 28.59 \\
        $f_{\text{gen}}^L$ & 8B & 54.1 & 0.89 & 28.59 \\
        $f_{\text{gen}}^G$ & 9B & 55.0 & 1.10 & 34.09 \\
        $f_{\text{link}}^M + f_{\text{gen}}^M$ & 7B & 61.3 & 1.24 & 57.18 \\
        $f_{\text{link}}^M + f_{\text{gen}}^L$ & 8B & 60.0 & 1.21 & 57.18 \\
        $f_{\text{link}}^M + f_{\text{gen}}^G$ & 9B & 60.8 & 1.28 & 62.68 \\
        $f_{\text{link}}^M \to \{ f_{\text{gen}}^M, f_{\text{gen}}^M \} \to f_{\text{msc}}^M$ & 7B & 64.0 & 1.42 & 114.36 \\
        $f_{\text{link}}^M \to \{ f_{\text{gen}}^L, f_{\text{gen}}^L \} \to f_{\text{msc}}^M$ & 8B & 63.5 & 1.50 & 114.36 \\
        $f_{\text{link}}^M \to \{ f_{\text{gen}}^G, f_{\text{gen}}^G \} \to f_{\text{msc}}^M$ & 9B & 62.7 & 1.53 & 125.36 \\
        $f_{\text{link}}^M \to \{ f_{\text{gen}}^M, f_{\text{gen}}^M , f_{\text{gen}}^M \} \to f_{\text{msc}}^M$ & 7B & 64.8 & 1.55 & 142.95 \\
        $f_{\text{link}}^M \to \{ f_{\text{gen}}^L, f_{\text{gen}}^L , f_{\text{gen}}^L \} \to f_{\text{msc}}^M$ & 8B & 64.1 & 1.54 & 142.95 \\
        $f_{\text{link}}^M \to \{ f_{\text{gen}}^G, f_{\text{gen}}^G , f_{\text{gen}}^G \} \to f_{\text{msc}}^M$ & 9B & 62.9 & 1.69 & 159.45 \\
        $f_{\text{link}}^M \to \{ f_{\text{gen}}^M, f_{\text{gen}}^L , f_{\text{gen}}^G \} \to f_{\text{msc}}^M$ & 9B & 65.6 & 1.68 & 148.45 \\
        CHESS+LLaMA-3 & 70B & 61.5 & NA & 1682.28 \\
        \bottomrule
    \end{tabular}
    \caption{Comparison of latency and FLOPs across different model configurations on the BIRD dataset.}
    \label{tab:latency}
\end{table}

\end{document}